\documentclass{article}

\usepackage[preprint]{neurips_2026}

\usepackage[utf8]{inputenc} % allow utf-8 input
\usepackage[T1]{fontenc}    % use 8-bit T1 fonts
\usepackage{hyperref}       % hyperlinks
\usepackage{url}            % simple URL typesetting
\usepackage{booktabs}       % professional-quality tables
\usepackage{amsfonts}       % blackboard math symbols
\usepackage{nicefrac}       % compact symbols for 1/2, etc.
\usepackage{microtype}      % microtypography
\usepackage[table]{xcolor}  % colors and table row/column coloring
\usepackage{wrapfig}
\usepackage{enumitem}
\usepackage{caption}
\usepackage{float}
\usepackage{colortbl}
\usepackage{makecell}
\usepackage[flushmargin]{footmisc}

\definecolor{cellteal}{RGB}{220, 239, 216}    % teal-50, "teacher endorses"
\definecolor{cellcoral}{RGB}{255, 215, 215}

\usepackage[most]{tcolorbox}
\tcbuselibrary{breakable}
\tcbuselibrary{skins, breakable}
\tcbuselibrary{skins, breakable, listings}
\definecolor{myblue}{RGB}{240, 246, 252}
\definecolor{myframe}{RGB}{180, 210, 235}
\definecolor{mytitle}{RGB}{200, 220, 238}
\usepackage{listings}
\definecolor{promptblue}{RGB}{14,31,145}
\definecolor{promptbg}{RGB}{232,233,248}
\lstdefinestyle{promptstyle}{
    basicstyle=\ttfamily\small,
    breaklines=true,
    columns=fullflexible,
    keepspaces=true,
    showstringspaces=false,
    frame=none,
    backgroundcolor=\color{promptbg},
    xleftmargin=0pt,
    xrightmargin=0pt,
    aboveskip=0pt,
    belowskip=0pt
}
\newtcolorbox{promptbox}[1]{
    enhanced,
    colback=promptbg,
    colframe=promptblue,
    boxrule=1pt,
    arc=3mm,
    outer arc=3mm,
    width=\textwidth,
    left=8pt,
    right=8pt,
    top=8pt,
    bottom=8pt,
    title=#1,
    coltitle=white,
    colbacktitle=promptblue,
    fonttitle=\bfseries,
    attach boxed title to top left={xshift=0mm,yshift=0mm},
    boxed title style={
        sharp corners,
        boxrule=0pt,
        colframe=promptblue,
        colback=promptblue,
        left=8pt,
        right=8pt,
        top=6pt,
        bottom=6pt
    }
}

\usepackage{titlesec}
\titlespacing*{\paragraph}{0pt}{3pt}{3pt}
\titlespacing*{\section}{0pt}{8pt}{4pt}
\titlespacing*{\subsection}{0pt}{6pt}{3pt}

\title{Beyond Outcome Rewards: Step-Level Self-Distilled Policy Optimization for Deep Search Agents}

\author{%
  \textbf{Haoze Wu}$^1$ \quad
  \textbf{Chuqiao Kuang}$^2$ \quad
  \textbf{Tianyi Zhuang}$^2$ \quad
  \textbf{Xiaoguang Li}$^2$ \\[4pt]
  $^1$The Hong Kong University of Science and Technology \quad
  $^2$Huawei Technologies Ltd. \\[2pt]
}

\begin{document}

\maketitle

\begin{abstract}
Deep search agents operate over trajectories spanning dozens of steps, yet standard reinforcement learning provides only a single outcome reward per trajectory — supervision that is far too sparse for effective credit assignment. 
On-policy self-distillation (OPSD) addresses this by using the model's own logits as dense token-level teachers, but extending it to search agents introduces a fundamental tension: the teacher, having access to privileged information such as the correct answer, produces a distribution that differs systematically from the student's exploration-based reasoning, and naive distillation causes the student to inherit this information asymmetry rather than learn better search strategies.
We resolve this tension through two contributions. 
First, we construct Evidence Anchors — concise, step-level evidence snippets extracted from the web — as privileged information that captures key reasoning steps without revealing the entire answer path. 
Second, we propose Step-Level Self-Distilled Policy Optimization (SSPO), which converts teacher–student disagreement into step-level advantage weights within GRPO, applied exclusively to incorrect trajectories. 
This design decouples what to update from how much to update: the outcome reward determines the direction of policy change, while the teacher modulates its magnitude at each step. Correct trajectories are left untouched, preserving their diversity.
On Qwen3-8B, SSPO consistently outperforms GRPO across BrowseComp, GAIA, and FRAMES, surpassing or matching GRPO trained with twice as many gradient steps while adding only about 5\% overhead per step from a single additional forward pass.\footnote{Code is available at \url{https://github.com/hkust-nlp/SSPO}. Correspondence to: \texttt{hwuds@connect.ust.hk}.}
\end{abstract}

\begin{figure}[H]
  \centering
  \begin{minipage}{0.495\textwidth}
    \centering
    \includegraphics[width=\linewidth]{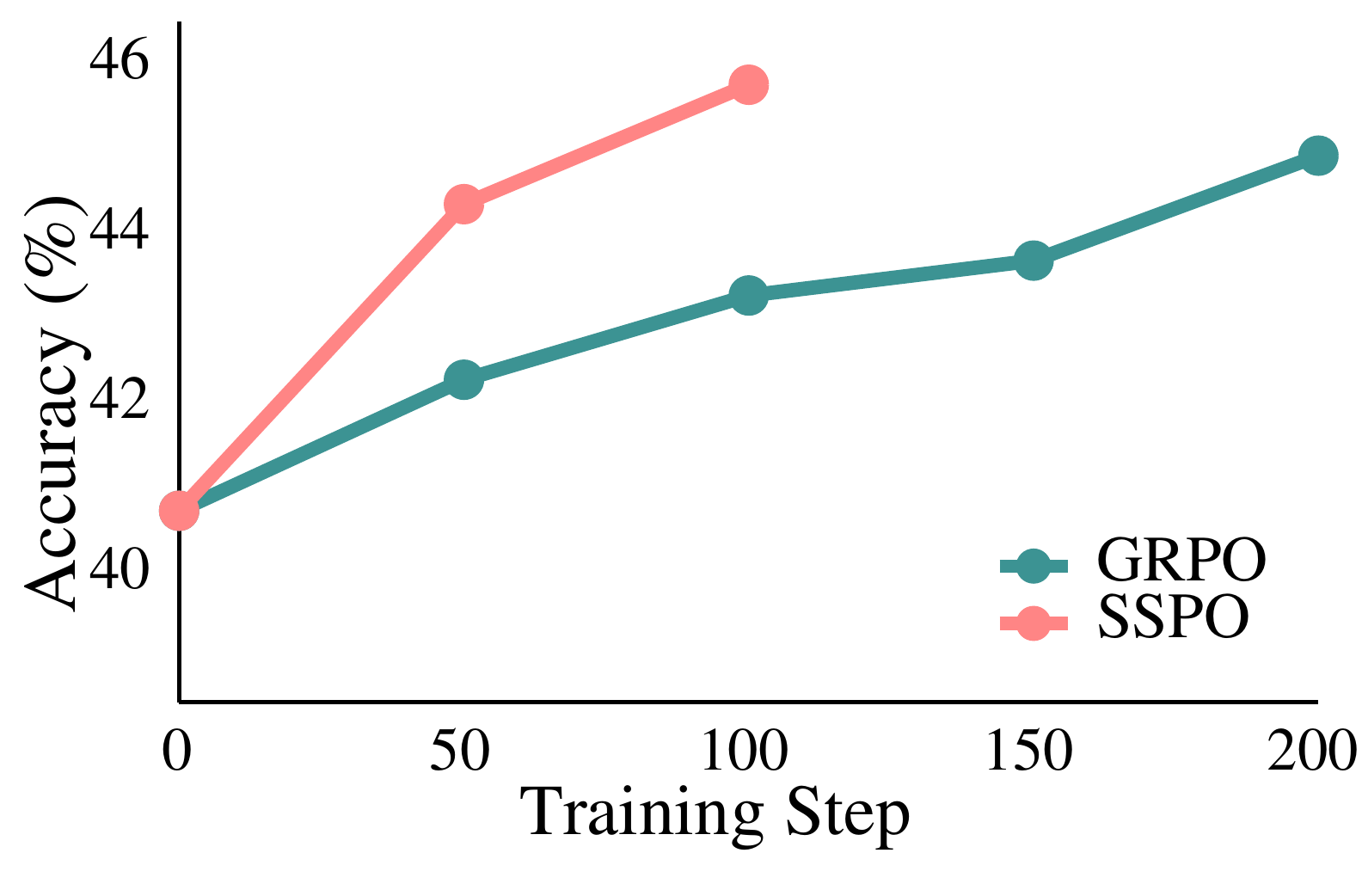}
  \end{minipage}
  \hfill  
  \begin{minipage}{0.495\textwidth}
    \centering
    \includegraphics[width=\linewidth]{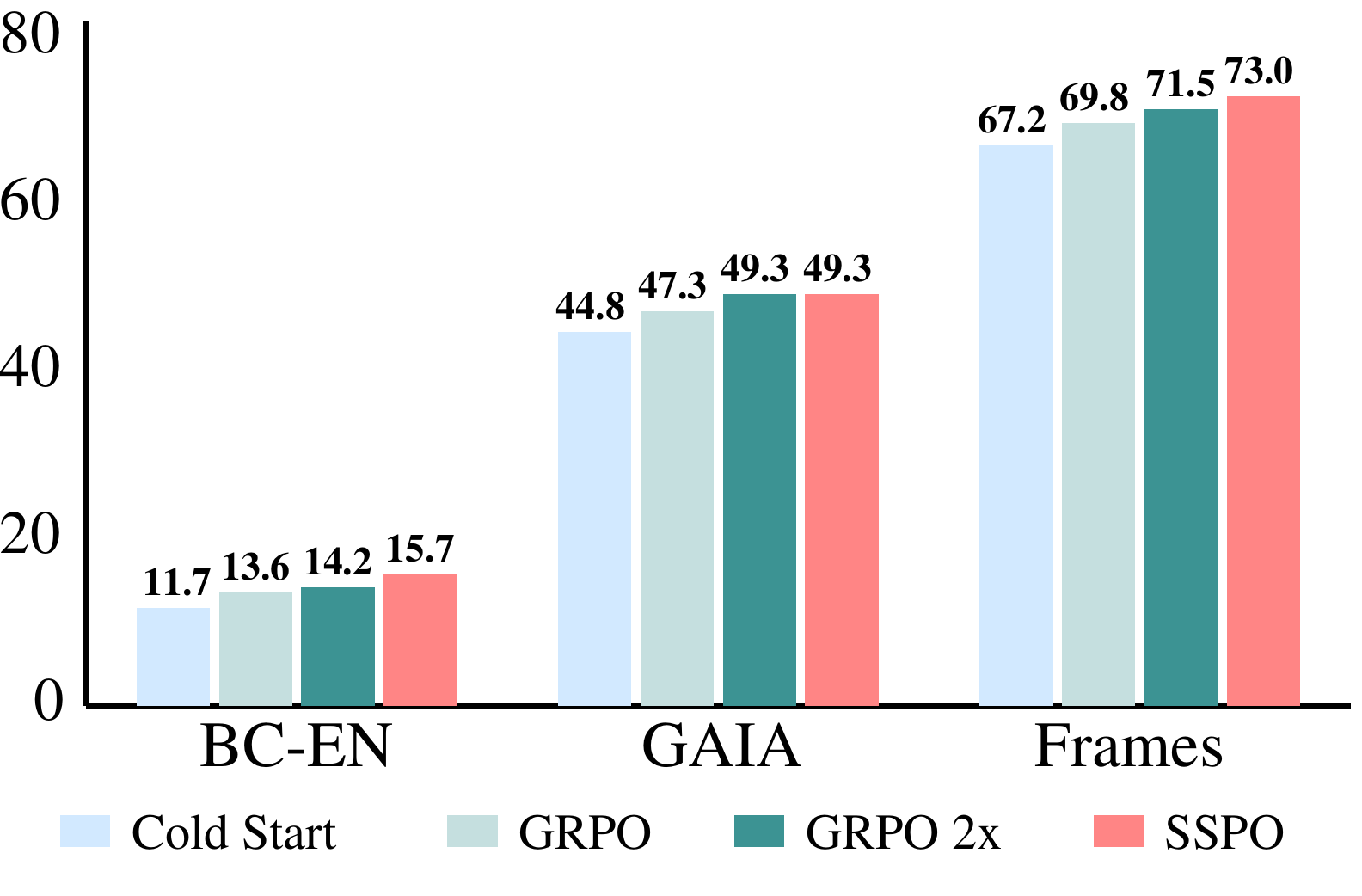}
  \end{minipage}
  \caption{\textbf{Comparison of performance across training methods. Left:} After introducing step-level self-distilled advantage weights (SSPO), performance improves substantially faster during training, surpassing GRPO trained for 200 steps after only 100 steps. \textbf{Right:} We report results for the cold-start model, GRPO after 100 and 200 steps, and SSPO after 100 steps.}
  \label{fig:overall}
\end{figure}

\section{Introduction}

Recent Large Language Model paradigms have rapidly evolved from generating appropriate responses to solving real-world tasks through tool use across domains such as coding, office productivity, finance, and research~\cite{li2026tooldecathlonbenchmarkinglanguage,gpt5-5,kimik26,opus47,glm51,gemini31pro,qwen36plus,mm27,deepseekai2026deepseekv4}. 
Among these applications, web search agents must answer vague, underspecified queries by actively exploring the open web — issuing searches, inspecting pages, and refining their reasoning over trajectories that often span dozens of steps~\cite{wei2025browsecompsimplechallengingbenchmark}. 
Training agents to perform such long-horizon search reliably requires post-training beyond Supervised Fine-Tuning (SFT), making reinforcement learning (RL) the dominant approach~\cite{liu2025webexplorer,li2025websailornavigatingsuperhumanreasoning,li2025websailorv2bridgingchasmproprietary,tao2025webshaperagenticallydatasynthesizing}. However, RL in this setting faces a fundamental challenge: trajectories containing 20+ steps receive only a single binary reward. With supervision this sparse, the model receives little guidance on which steps led to success or failure, making credit assignment the central bottleneck for deep search agents.

A natural way to address sparse rewards is to introduce denser supervision. SFT provides token-level feedback through teacher trajectories, but its Off-Policy nature introduces exposure bias and limits generalization to unseen queries~\cite{schmidt-2019-generalization,pmlr-v267-chu25c,wu2026generalizationsftreinforcementlearning,shenfeld2026selfdistillationenablescontinuallearning}. On-Policy Distillation (OPD)~\cite{lu2025onpolicydistillation,song2026surveyonpolicydistillationlarge,li2026rethinking} combines both benefits: the student performs on-policy rollouts while a teacher model provides token-level scoring. More recent work extends this to self-distillation, where the teacher is constructed from the student itself — using privileged information such as reference solutions or environment feedback as a prefix~\cite{zhao2026selfdistilledreasoneronpolicyselfdistillation,yang2026selfdistilledrlvr,hubotter2026reinforcement,shenfeld2026selfdistillationenablescontinuallearning,sang2026crispcompressedreasoningiterative,li2026unifyinggrouprelativeselfdistillationpolicy}, eliminating the need for a separate teacher model.

Despite this progress, existing self-distillation methods have been developed and evaluated primarily on single-turn reasoning tasks such as math and code. Two properties make self-distillation feasible in those settings: the gap between teacher and student behavior is modest, and a natural form of privileged information — reference solutions or execution feedback — already exists. Extending self-distillation to multi-turn search agents breaks both assumptions. The information asymmetry becomes extreme: the teacher, seeing curated evidence and the answer, collapses to 3.5 tool calls per trajectory, while the student, navigating the open web, requires 17.7 (Table~\ref{tab:leak}). Directly distilling this collapsed distribution causes the student to mimic the teacher’s brevity; when privileged information is unavailable, it abandons tool use prematurely rather than learning better search strategies. An equally serious problem compounds the first: unlike math or code, open-ended web search has no ready-made privileged information. A form of supervision that is both informative to the teacher and safe to distill to the student does not naturally exist, and constructing one is itself an open challenge.

Our key insight is that search supervision is naturally defined at the level of information-seeking actions rather than token generation. We therefore align privileged supervision with the structure of search through two design choices. First, we construct \textbf{Evidence Anchors}: compact, step-level evidence snippets extracted from the web that capture information needed to answer a question without revealing the answer path. Each question is associated with a small set of such anchors, serving as prefixes for the self-teacher. Second, we propose \textbf{Step-Level Self-Distilled Policy Optimization} (SSPO). Rather than minimizing divergence between teacher and student distributions, SSPO converts teacher–student disagreement into step-level advantage weights within GRPO~\cite{shao2024deepseekmathpushinglimitsmathematical,Guo_2025}, applied only to incorrect trajectories. This design decouples update direction from update magnitude: the outcome reward determines whether to reinforce or suppress a trajectory, while the teacher modulates how much each step contributes. Correct trajectories are left untouched, preserving behavioral diversity.

We evaluate SSPO on Qwen3-8B~\cite{yang2025qwen3technicalreport} across three information-seeking benchmarks — BrowseComp~\cite{wei2025browsecompsimplechallengingbenchmark}, GAIA~\cite{mialon2023gaiabenchmarkgeneralai}, and FRAMES~\cite{krishna2024factfetchreasonunified}. SSPO consistently outperforms GRPO across all benchmarks, and notably surpasses GRPO trained with twice as many gradient steps while adding only about 5\% computational overhead per training step. Our ablation studies further validate two key design decisions. First, directly matching the teacher distribution collapses the student's tool use and underperforms even GRPO, confirming that decoupling update magnitude from direction is essential. Second, step-level advantage weights consistently outperform token-level counterparts, whose fine-grained signals are misaligned with the natural unit of information-seeking actions.

In summary, our contributions are:
\textbf{(a)} We show that extending self-distillation to multi-turn search breaks the assumptions underlying single-turn self-distillation, requiring a fundamentally different design.
\textbf{(b)} We introduce Evidence Anchors: compact, step-level evidence snippets as privileged information for open-ended web search, aligned with the natural unit of information-seeking actions.
\textbf{(c)} We propose SSPO, which uses step-level self-distilled signals as advantage weights rather than optimization targets, achieving superior sample efficiency over GRPO.

\section{Preliminaries}

\subsection{Search Agent}

We build our search agent based on the ReAct paradigm~\cite{yao2023reactsynergizingreasoningacting,xie2025logicrlunleashingllmreasoning}, where the model interleaves reasoning and actions until producing the final answer.
Our action space includes two tools connected to the real web: \texttt{search} and \texttt{browse}, their details could be found in Appendix~\ref{app:tool-schema}.
After receiving a user query, the agent attempts to provide an accurate answer through multiple \textit{Thought Action Observation} steps.
An agent with $T$ steps can be formulated as follows:
\begin{equation*}
H_T = \{\mathtt{q}_{user}, \mathtt{t}_1, \mathtt{a}_1, \mathtt{o}_1, \ldots, \mathtt{t}_{T-1}, \mathtt{a}_{T-1}, \mathtt{o}_{T-1}, \mathtt{t}_T\}
\end{equation*}
At time step $\tau$, the agent generates a thought $\mathtt{t}_\tau$ and a tool call $\mathtt{a}_\tau$ (if it is not the final step) conditioned on the full interaction history, i.e., $\mathtt{t}_\tau, \mathtt{a}_\tau \sim \pi(\mathtt{t}, \mathtt{a} \mid H_{\tau-1})$.
The output of the tool is treated as the observation $\mathtt{o}_\tau$ for this step.

\subsection{On-Policy Learning}

\paragraph{Agentic Reinforcement Learning.}
Deep search, similar to math and logic tasks~\cite{shao2024deepseekmathpushinglimitsmathematical,xie2025logicrlunleashingllmreasoning}, is typically formulated under the Reinforcement Learning with Verifiable Rewards (RLVR) setting.
In this setting, a correctness reward can be assigned by evaluating whether the agent’s prediction is semantically consistent with the ground truth: $R_{\text{correct}} = \text{\texttt{is\_equal(pred, ground\_truth)}} \in \{0,1\}$.
In addition, training often incorporates a format reward $R_{\text{format}}$ to encourage adherence to the ReAct paradigm~\cite{yao2023reactsynergizingreasoningacting}.
The final training reward is defined as:
\begin{equation}
  R_{final} = R_{correct} + 0.2 \times R_{format}
  \label{eq:reward}
\end{equation}
Similar to other works~\cite{liu2025webexplorer,li2025websailornavigatingsuperhumanreasoning,li2025websailorv2bridgingchasmproprietary}, we use the Group Relative Policy Optimization (GRPO) algorithm to update the model~\cite{shao2024deepseekmathpushinglimitsmathematical,Guo_2025}.
It addresses the challenge of estimating baselines in policy gradient methods by comparing responses within a group. 
Specifically, for each question $q$, GRPO samples $G$ responses $\{y^{(1)}, \ldots, y^{(G)}\}$ from the current policy and normalizes the rewards to obtain a sequence-level advantage estimate:
\begin{equation}
    A^{(i)} = \frac{R_{final}(q, y^{(i)}) - \mu_G}{\sigma_G},
\end{equation}
where $\mu_G$ and $\sigma_G$ denote the mean and standard deviation of rewards within the group, respectively.
The policy is then updated by maximizing a clipped surrogate objective:
\begin{equation}
    \mathcal{L}_{\text{GRPO}}(\theta) = \mathbb{E}_{q\sim\mathcal{D}}\left[
        \frac{1}{G} \sum_{i=1}^{G} \frac{1}{|y^{(i)}|}
        \sum_{t=1}^{|y^{(i)}|}
        \min\!\left(
            \rho_t^{(i)} A^{(i)}_t,\ 
            \text{clip}\!\left(\rho_t^{(i)}, 1-\epsilon, 1+\epsilon\right) A^{(i)}_t
        \right)
    \right]
    \label{eq:grpo}
\end{equation}
where $\rho_t^{(i)} = \pi_\theta(y_t^{(i)} \mid q) / \pi_{\theta_{\text{old}}}(y_t^{(i)} \mid q)$ is the per-token importance sampling ratio, and $\epsilon$ is the clipping threshold that constrains the policy update to a trust region.
All tokens in the trajectory share the sequence-level advantage $A_t^{(i)}=A^{(i)}$.
To improve training efficiency, we adopt the duplication strategy proposed in WebSailor~\cite{li2025websailornavigatingsuperhumanreasoning}, where groups with non-zero variance within the batch are duplicated to replace zero-variance groups, as the latter provide no useful training signal.

\paragraph{On-Policy Self Distillation (OPSD).}
Although GRPO has become the default algorithm for training agents, it has a key limitation in settings with long reasoning chains: relying solely on outcomes lacks process-level supervision. This is especially problematic for incorrect responses, where a small error in the final few tokens may lead to an incorrect answer, yet all preceding tokens receive the same level of penalty.
Recent works, including OPSD~\cite{yang2026selfdistilledrlvr,zhao2026selfdistilledreasoneronpolicyselfdistillation,hubotter2026reinforcement,shenfeld2026selfdistillationenablescontinuallearning,ye2026onpolicycontextdistillationlanguage,sang2026crispcompressedreasoningiterative,li2026unifyinggrouprelativeselfdistillationpolicy}, attempt to alleviate the lack of process-level supervision in long CoT~\cite{wei2023chainofthoughtpromptingelicitsreasoning} without relying on other teacher models~\cite{lu2025onpolicydistillation}.
These methods provide the model with privileged information $c_{privileged}$ as a form of self-teaching, such as reference solutions, environment feedback, or better rollouts, thereby enabling token-level supervision for the student. The training objective is formulated as follows:
\begin{equation}
  \mathcal{L}_{\text{OPSD}}(\theta) = \mathbb{E}_{q\sim\mathcal{D},y\sim P_S(\cdot\mid q)}\left[\frac{1}{|y|}\sum_{t=1}^{|y|}\mathcal{D}(P_T\mid\mid P_S)\right]
  \label{eq:opsd}
\end{equation}
where $\mathcal{D}$ denotes a divergence measure (e.g., KL divergence), and $\theta$ represents the parameters of the student model.
The teacher and student distributions can be expressed as:
\begin{align}
\text{Teacher}: P_T(y\mid q,y_{<t}) = \pi_{\hat{\theta}}(\cdot\mid q,y_{<t},c_{privileged})\,\,
\text{Student}: P_S(y\mid q,y_{<t}) = \pi_{\theta}(\cdot\mid q,y_{<t})
\end{align}
For training stability, the teacher model’s parameters $\hat{\theta}$ are typically not updated through gradient training; common choices include keeping them fixed, updating them via EMA, or synchronizing them with the student model at regular step intervals.

Although significant progress has been made, existing work is largely limited to relatively simple single-turn reasoning tasks.
Our work explores how to leverage self-distillation signals to provide more fine-grained supervision for training long-horizon search agents.

\section{Step-Level Self-Distilled Policy Optimization}

Deep search agents operate through information-seeking actions whose utility emerges only at the level of complete search steps. A retrieval action is valuable not because of individual token choices, but because it effectively localizes uncertainty, gathers relevant evidence, and shapes subsequent exploration. Motivated by this observation, both our privileged information and supervision are aligned with the structure of search steps.

\subsection{Privileged Information: Evidence Anchors}

In single-turn settings, constructing privileged information is relatively straightforward.
For mathematical tasks, many open-source datasets provide reference solutions~\cite{zhao2026selfdistilledreasoneronpolicyselfdistillation,guha2025openthoughtsdatarecipesreasoning}; for tool use and coding, error messages can be directly obtained~\cite{hubotter2026reinforcement}; even in the absence of explicit feedback, the highest-scoring rollout among multiple candidates can be selected as privileged teacher information~\cite{hubotter2026reinforcement}.
However, for BrowseComp-style tasks~\cite{wei2025browsecompsimplechallengingbenchmark}, models often require context windows spanning tens or even hundreds of thousands of tokens.
In such settings, providing the best rollout as privileged information is impractical, while supplying only the final answer is insufficient to effectively guide multi-step reasoning and action.
What is needed is privileged information structured around individual search actions — capable of signaling whether each step retrieves the right evidence.

\begin{figure}[h]
  \centering
  \includegraphics[width=\textwidth]{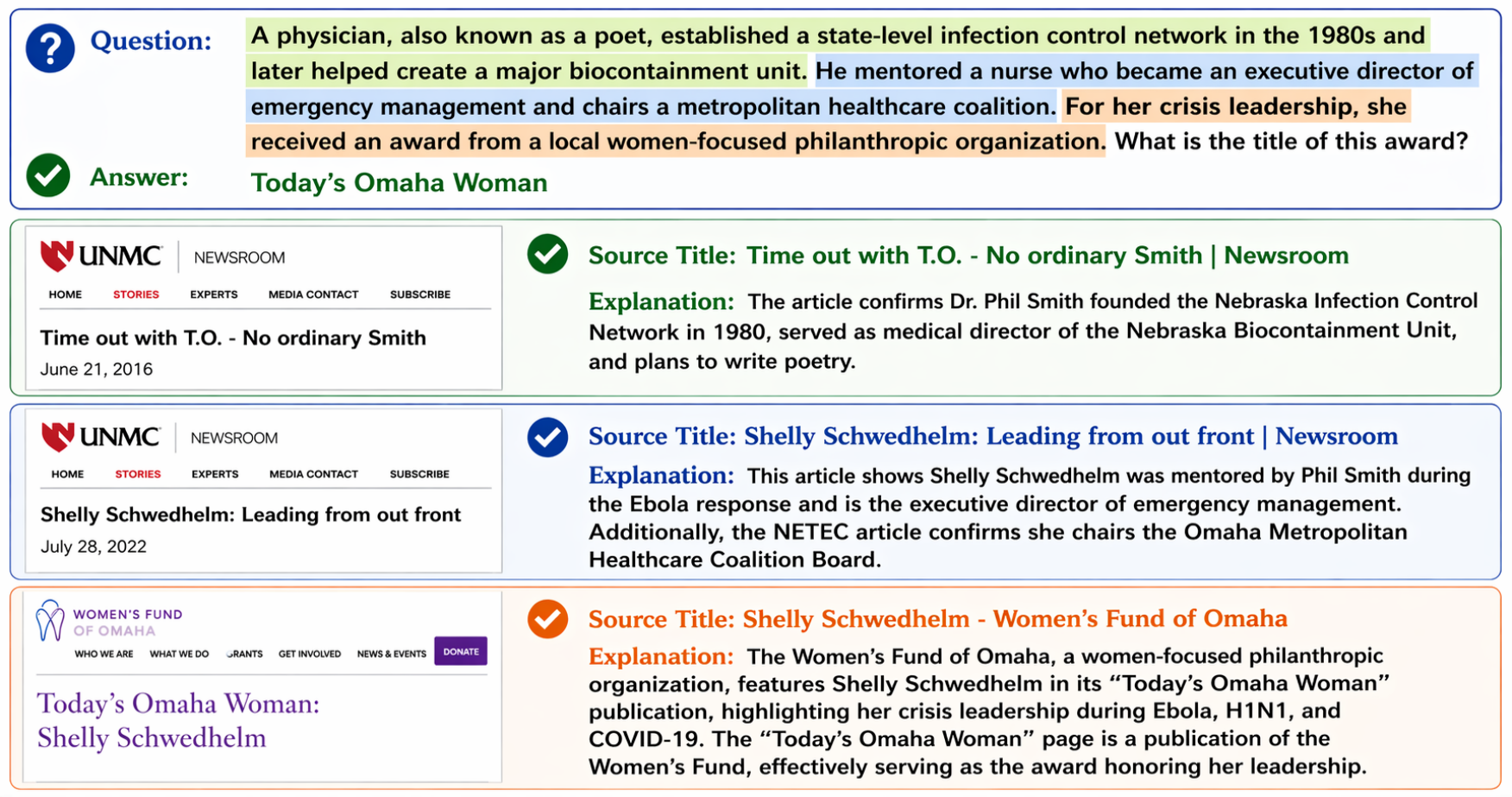}
  \caption{An example question with three evidence anchors. The differently colored evidence anchors correspond to the highlighted conditions in the question.}
  \label{fig:example_anchors}
\end{figure}

To construct such privileged information, we construct key pieces of evidence for each QA pair as \textbf{Evidence Anchors}.
Specifically, we prompt a SOTA LLM to identify as many pieces of evidence as possible that support the ground-truth answer for each question.
The prompt we use can be seen in Figure~\ref{fig:evidence-anchors-prompt-template}.
We collect evidence anchors for over 6,000 QA pairs, with an average of 5.24 anchors per question.
Figure~\ref{fig:example_anchors} presents an example from our training data.
Detailed statistics and correctness validation are provided in Appendix~\ref{app:evidence-anchors}.
These privileged signals are incorporated into the teacher model’s prompt only for incorrect trajectories, as illustrated in Figure~\ref{fig:teacher-prompt-template}.
In practice, in addition to evidence anchors, we explicitly provide the teacher with the incorrect answers generated from the student’s rollouts, preventing the teacher from repeating the same mistakes in its output distribution.

\begin{figure*}[!ht]
\centering
\begin{minipage}{\textwidth}
\begin{promptbox}{}
\begin{lstlisting}[style=promptstyle]
Question: {question}
You previously attempted to obtain the answer {rollout answer}, but it is incorrect. 
The following are some evidence anchors that may help you:
- {source title}: {explanation} 
...
[IMPORTANT]: Do not answer directly from the anchors.
Correctly solve the original question:
\end{lstlisting}
\end{promptbox}
\end{minipage}
\caption{Teacher prompt template incorporating evidence anchors and previously generated incorrect answers from the student model.}
\label{fig:teacher-prompt-template}
\end{figure*}

\subsection{Step-Level Self-Distilled Advantage Weights}
\label{sec:ssaw}

\begin{wraptable}{r}{0.35\linewidth}
\small
\centering
\begin{tabular}{l|cc}
\toprule
 & \textbf{Pass@1} & \textbf{\#Turn} \\
\midrule
Question   & 46 & 17.7 \\
+ Anchors  & 82 & 3.5 \\
+ Answer   & 96 & 2.4 \\
\bottomrule
\end{tabular}
\caption{Comparison of accuracy and average steps before and after providing privileged information.}
\label{tab:leak}
\end{wraptable}

As discussed in prior work~\cite{yang2026selfdistilledrlvr}, directly optimizing Equation~\ref{eq:opsd} can lead to privileged information leakage, which may degrade model performance in the later stages of training.
This issue becomes more pronounced in multi-turn settings, where shortcuts introduced by privileged information affect not only the reasoning process itself but also significantly reduce the number of tool calls required for information gathering.
To quantify this effect, we sample 50 examples from the training set and compare the cold-start model’s accuracy and trajectory length under different information conditions.
As shown in Table~\ref{tab:leak}, providing evidence anchors in addition to the question improves accuracy from 46 to 82, while reducing the average number of steps from 17.7 to 3.5.
When the final answer is further revealed, accuracy increases to 96, and the average number of steps decreases to 2.4.
This substantial discrepancy suggests that the teacher's distribution with privileged information is not an appropriate optimization target for the student.
Instead, using self-distillation signals as weighting factors, rather than as gradient directions, helps mitigate this issue~\cite{yang2026selfdistilledrlvr}. 
Under this formulation, the update direction remains determined by environment rewards, while the teacher’s privileged distribution only modulates the update magnitude, preventing the student from directly fitting an information-asymmetric target.
Beyond privileged-information leakage, SRPO~\cite{li2026unifyinggrouprelativeselfdistillationpolicy} further shows that applying OPSD signals only to incorrect trajectories, while preserving the original GRPO objective for correct ones, yields better performance.
Applying distillation to already-correct trajectories introduces optimization ambiguity: the model is pushed toward a specific teacher distribution despite having already solved the task, unnecessarily suppressing the diversity of correct solutions.
Together, these two insights motivate our method design: we use self-distillation signals as step-level advantage weights rather than optimization targets, and restrict their application to incorrect search-agent trajectories.

For each student-generated step $\tau$ in trajectory $y^{(i)}$, its thinking tokens $\mathtt{t}_\tau$ and tool-calling tokens $\mathtt{a}_\tau$ jointly constitute one action in the agentic MDP~\cite{zhang2026the}. Under autoregressive factorization, the teacher and student assign the following conditional joint probabilities to this complete step:
\begin{equation}
\small
P_T(\mathtt{t}_\tau,\mathtt{a}_\tau)
= \prod_{k\in\tau}\pi_{\hat{\theta}}(y_k\mid c_{\text{privileged}},q,y_{<k}),
\qquad
P_S(\mathtt{t}_\tau,\mathtt{a}_\tau)
= \prod_{k\in\tau}\pi_{\theta}(y_k\mid q,y_{<k}).
\label{eq:step-joint-prob}
\end{equation}
Equation~\ref{eq:step-joint-prob} follows standard autoregressive factorization, under which the joint probability of a step is the product of its token-level conditional probabilities. We compute both likelihoods by teacher-forcing the same student-generated step under the corresponding teacher and student configurations: $P_T$ uses the privileged context and teacher parameters, whereas $P_S$ uses the original context and current policy parameters.

We then define the privileged-information gain as the log joint-likelihood ratio of this step action:
\begin{equation}
\Delta_\tau^{step}
= \text{\texttt{sg}}\left(
\log\frac{P_T(\mathtt{t}_\tau,\mathtt{a}_\tau)}
{P_S(\mathtt{t}_\tau,\mathtt{a}_\tau)}
\right),
\label{eq:step-gain}
\end{equation}
where \texttt{sg} denotes stop-gradient.
For tokens $y_t$ belonging to step $\tau$ ($t \in \tau$), they share the same step-level advantage weight:
\begin{equation}
w_t \overset{t \in \tau}{=} w_{\tau}^{step} = \min(\exp(\text{sign}(A^{(i)})\cdot\Delta_\tau^{step}),1+\epsilon)
\label{eq:step-level-weight}
\end{equation}
where $\epsilon$ is a positive hyperparameter, and the $\min$ operation prevents extreme gradient magnitudes.
It is important to clarify that trajectories with $R_{final} < 1$ in Equation~\ref{eq:reward} are treated as incorrect trajectories, since they fail to produce a factually equivalent final answer\footnote{An incorrect trajectory may still have a positive group-relative advantage, e.g., when it receives the format reward and outperforms other rollouts in the same group. Hence, Equation~\ref{eq:step-level-weight} explicitly considers the advantage sign.}. Finally, we replace the advantage term in Equation~\ref{eq:grpo} with $\hat{A}^{(i)}_t = w_tA_t^{(i)}\text{ \texttt{if} }R_{final}<1\text{ \texttt{else} }A^{(i)}$.

\begin{wraptable}{r}{0.5\textwidth}
\centering
\small
\scalebox{0.92}{
\begin{tabular}{c|cc}
\toprule
 & $\Delta^{\text{step}}_\tau > 0$
 & $\Delta^{\text{step}}_\tau < 0$ \\
\midrule
$A^{(i)} > 0$
 & \cellcolor{cellteal}\shortstack{$w > 1$ \\ Reward amplified}
 & \cellcolor{cellcoral}\shortstack{$w < 1$ \\ Reward dampened} \\
$A^{(i)} < 0$
 & \cellcolor{cellteal}\shortstack{$w < 1$ \\ Penalty reduced}
 & \cellcolor{cellcoral}\shortstack{$w > 1$ \\ Penalty amplified} \\
\bottomrule
\end{tabular}}
\caption{Step-level advantage weight modulation.}
\label{tab:advantage_cases}
\end{wraptable}

As summarized in Table~\ref{tab:advantage_cases}, the weight $w$ modulates each token's contribution along two axes: the sign of the trajectory-level advantage $A^{(i)}$ and the step-level teacher--student agreement $\Delta^{\text{step}}_\tau$.
For incorrect trajectories with $A^{(i)} < 0$, which represent the typical case for failed rollouts, we use the weight $\min(P_S / P_T,\, 1 + \epsilon)$. When the student is overconfident on a step rejected by the teacher ($P_S \to 1,\, P_T \to 0$, i.e., $\Delta^{\text{step}}_\tau < 0$), $w \in (1,\, 1 + \epsilon)$ amplifies the penalty, discouraging repeated mistakes. Conversely, when the student is uncertain on a step endorsed by the teacher ($P_S \to 0,\, P_T \to 1$, i.e., $\Delta^{\text{step}}_\tau > 0$), $w \in (0,\, 1)$ reduces the penalty, preserving valid intermediate reasoning within failed rollouts.
For incorrect trajectories with $A^{(i)} > 0$, which may occur on challenging questions when a trajectory receives format rewards despite producing an incorrect final answer, we adopt the symmetric weight $\min(P_T / P_S,\, 1 + \epsilon)$: a confident teacher paired with an uncertain student amplifies the reward, while a confident student paired with an uncertain teacher dampens it.
Across all cases, SSPO adjusts the \emph{magnitude} of updates at step-level resolution without altering the trajectory-level update \emph{direction}.

\section{Experiment}
\label{sec:exp}

\subsection{Experimental Setup}

\paragraph{Training Data.}
Before On-Policy learning, we first followed the WebExplorer~\cite{liu2025webexplorer} pipeline to collect approximately 4,000 English reasoning trajectories with correct final answers and without obvious tool-calling errors, which we used for cold-start initialization.
Further details of these trajectories are provided in Appendix~\ref{app:cold-start}.
For On-Policy learning, we sampled around 6,000 English QA pairs from the open-source DeepForge dataset~\cite{zhou2026offseekeronlinereinforcementlearning}, using a difficulty ratio of 1.5:3.5:3.5:1.5 across the four levels. We further employed DeepSeek-V3.2~\cite{deepseekai2025deepseekv32pushingfrontieropen} to construct evidence anchors for each QA pair.

\paragraph{Benchmarks.}
BrowseComp~\cite{wei2025browsecompsimplechallengingbenchmark} is a highly challenging information-retrieval benchmark introduced by OpenAI. GAIA~\cite{mialon2023gaiabenchmarkgeneralai} is a widely used benchmark for general AI assistants; following WebThinker~\cite{li2025webthinkerempoweringlargereasoning}, we evaluate on its text-only subset. FRAMES~\cite{krishna2024factfetchreasonunified}, introduced by Google, is used to assess factual accuracy and reasoning capability.
Due to API cost constraints, we evaluate intermediate checkpoints on subsets of BrowseComp and FRAMES, denoted as \{BC, FRAMES\}-Sub.
We adopt the Avg@4 metric to reduce variance, with the temperature set to 0.6 and top-$p$ set to 0.95.

More experimental details could be found in Appendix~\ref{app:more-exp-detail}.

\subsection{Experimental Results}

\paragraph{Main Results.}
Table~\ref{tab:main-res} presents the performance of representative small-scale search agents, alongside our models trained with cold-start initialization, GRPO, and SSPO, as well as the average number of turns required to solve each problem.
Despite differences in search scaffolding and the fact that existing agents of comparable size are typically trained on substantially larger datasets, our models achieve competitive—often superior—performance, validating the effectiveness of our training pipeline. Our primary focus, however, is on the performance gains that SSPO brings over GRPO within the same scaffold.
Models trained with SSPO consistently outperform their GRPO counterparts by a significant margin. Using the average score across three benchmarks as a representative metric, GRPO achieves a +2.4 improvement over the cold-start baseline, whereas SSPO delivers a markedly larger gain of +4.8.
Importantly, these improvements do not come at the expense of reasoning efficiency. The SSPO-trained model requires only a marginally higher number of turns compared to GRPO (20.5 vs. 20.0), indicating that its performance gains are achieved with minimal additional computational cost.
As shown in Figure~\ref{fig:overall} and Figure~\ref{fig:main-res}, SSPO demonstrates substantially higher learning efficiency, consistently outperforming GRPO across all three benchmarks at the same number of training steps. 
To further examine the impact of process supervision, we double the number of training steps for GRPO; notably, SSPO trained for 100 steps already surpasses GRPO trained for 200 steps, highlighting the superior sample efficiency enabled by process-level supervision.
Beyond benchmark scores, these gains come with negligible training cost — only about 5\% additional overhead from teacher scoring; detailed wall-clock and compute statistics are provided in Appendix~\ref{app:more-exp-res}.

\begin{table}[!ht]
    \centering
    \setlength{\tabcolsep}{6pt}
    \renewcommand{\arraystretch}{1.15}
    \scalebox{0.96}{
    \begin{tabular}{lcccc}
    \toprule
     \textbf{Model}
     & \textbf{BrowseComp}~\cite{wei2025browsecompsimplechallengingbenchmark}
     & \textbf{GAIA}~\cite{mialon2023gaiabenchmarkgeneralai}
     & \textbf{Frames}~\cite{krishna2024factfetchreasonunified}
     & \textbf{Average} \\
    \midrule
    \rowcolor{gray!12}
    \multicolumn{5}{c}{\small\textit{Small Size Search Agent (<10B)}} \\
    \midrule
    \textbf{WebSailor-7B}~\cite{li2025websailornavigatingsuperhumanreasoning} & 6.7  & 37.9 & -- & -- \\
    \textbf{MiroThinker-8B-DPO-v0.1}~\cite{miromind2025mirothinker} & 8.7  & 46.6 & 64.4 & 39.9 \\
    \textbf{AFM-WebAgent-7B} (RL)~\cite{li2025chainofagentsendtoendagentfoundation} & 5.8 & 40.8 & -- & --\\
    \textbf{DeepDive-9B} (RL)~\cite{lu2025deepdiveadvancingdeepsearch} & 6.3  & -- & -- & -- \\
    \textbf{WebExplorer-8B} (RL)~\cite{liu2025webexplorer}           & \underline{15.7} & 50.0 & \underline{75.7} & \underline{47.1} \\
    \textbf{OffSeeker-8B} (DPO)~\cite{zhou2026offseekeronlinereinforcementlearning} & 12.8 & \underline{51.5} & --   & --   \\
    \midrule
    \rowcolor{gray!12}
    \multicolumn{5}{c}{\small\textit{Our Models (based on Qwen3-8B)}} \\
    \midrule
    \textbf{Our-8B} (Cold-Start) & 11.7$_{(37.7)}$ & 44.8$_{(14.9)}$ & 67.2$_{(7.2)}$ & 41.2$_{(19.9)}$ \\
    \textbf{Our-8B} (GRPO)       & 13.6$_{(38.0)}$ & 47.3$_{(14.5)}$ & 69.8$_{(7.6)}$ & 43.6$_{(20.0)}$ \\
    \textbf{Our-8B} (SSPO)       & \underline{\textbf{15.7}$_{(39.4)}$} & \textbf{49.3}$_{(14.5)}$ & \textbf{73.0}$_{(7.7)}$ & \textbf{46.0}$_{(20.5)}$ \\
    \bottomrule
    \end{tabular}}
    \caption{\textbf{Main results on the BrowseComp, GAIA, and Frames benchmarks.}
    Subscripts indicate the average number of search turns required per problem.
    \underline{Underlined} values denote the best overall performance in each column, while \textbf{bold} values highlight the best results among our models.
    Results for other search agents are reported from their respective original publications.}
    \label{tab:main-res}
\end{table}

\begin{figure}[H]
  \centering
  \includegraphics[width=\textwidth]{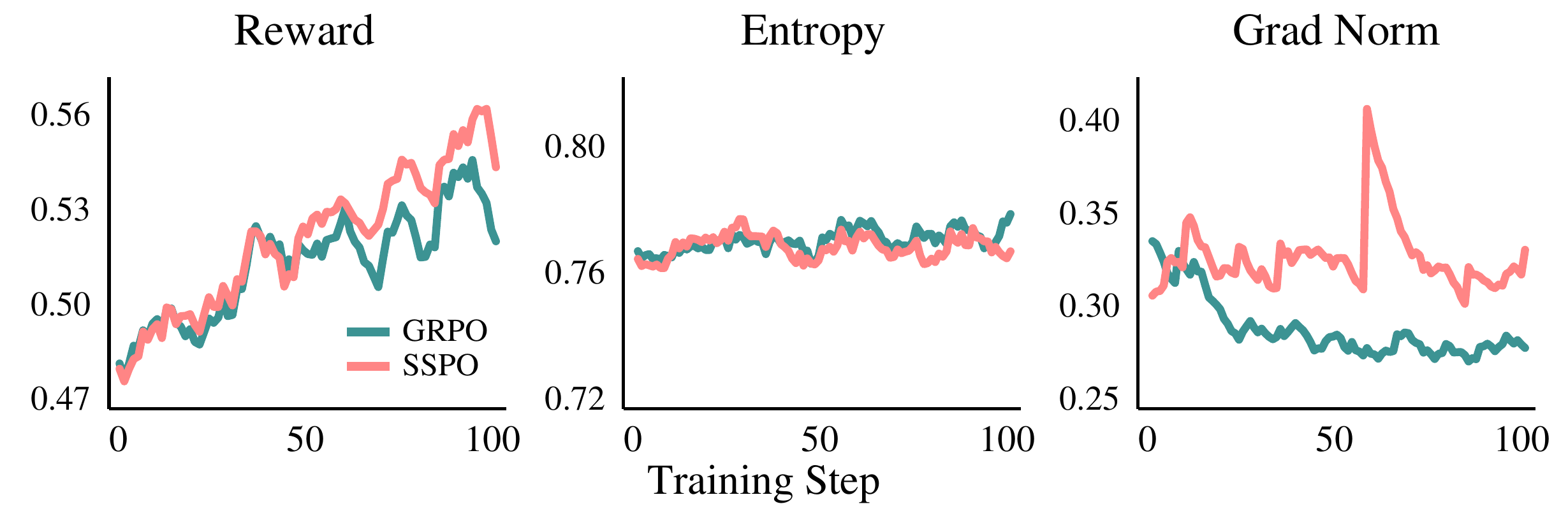}
  \caption{Comparison of training dynamics between GRPO and SSPO (EMA Smoothed).}
  \label{fig:dynamics}
\end{figure}

\paragraph{Training Dynamics.}
Beyond benchmark scores, we analyze the training dynamics in Figure~\ref{fig:dynamics}.
After an initial warm-up phase of approximately 30--50 steps, SSPO achieves higher rewards than GRPO and maintains this advantage throughout training, indicating more effective optimization in our experiments.
Notably, the entropy curves of both methods remain closely aligned, suggesting that the performance gains of SSPO are unlikely to stem primarily from increased exploration, but rather from more effective policy updates.
This observation is further supported by the gradient norm curves: GRPO exhibits a steadily decreasing gradient norm, indicating early saturation of policy updates, whereas SSPO maintains a relatively higher gradient norm throughout training, reflecting more sustained learning signals.
Although SSPO shows a transient spike in gradient norm around step 50, this spike quickly subsides and does not appear to harm training stability, as evidenced by the continued improvement in reward.
Overall, these results suggest that SSPO improves upon GRPO mainly by enabling more sustained and effective parameter updates, rather than by substantially altering the exploration--exploitation balance.

\subsection{Ablation Study}
\label{sec:why-step-level}

We organize the ablation around three questions: how direct distillation compares with advantage weighting, whether the weighting signal should be defined at token or step level, and what roles the two privileged-information inputs play. Unless otherwise stated, all results are measured on BC-Sub after 50 training steps.

\begin{table}[H]
\centering
\small
\setlength{\tabcolsep}{5pt}
\resizebox{\textwidth}{!}{%
\begin{tabular}{l l c c c c}
  \toprule
  \textbf{Method} & \textbf{Use of self-distillation signal} & \textbf{Granularity} & \textbf{Evidence Anchors} & \textbf{BC-Sub} & \textbf{Avg. turns} \\
  \midrule
  GRPO & None; outcome reward only & -- & No & 12.8 & 36.1 \\
  OPSD adaptation & Direct optimization target & Token & Yes & 11.8 & 30.9 \\
  RLSD/SRPO adaptation & Advantage weighting & Token & Yes & 11.8 & 35.5 \\
  \textbf{Ours (SSPO)} & \textbf{Advantage weighting} & \textbf{Step} & Yes & \textbf{14.5} & 37.5 \\
  \bottomrule
\end{tabular}}
\caption{Comparison of the outcome-only baseline and self-distillation variants after 50 training steps. The RLSD/SRPO adaptation uses the self-distillation signal as token-level advantage weights, following RLSD, and applies it only to incorrect trajectories, following SRPO.}
\label{tab:step-vs-token}
\end{table}

\paragraph{Direct Distillation vs. Advantage Weighting.}
GRPO provides the outcome-only baseline, reaching 12.8 accuracy with an average of 36.1 turns. Under the same training budget, the direct OPSD adaptation (Appendix~\ref{app:opsd-obj}) obtains lower accuracy than GRPO (11.8 vs. 12.8) and produces substantially shorter trajectories (30.9 vs. 36.1 turns). Together with Table~\ref{tab:leak}, this pattern is consistent with privileged-information shortcutting: the teacher can solve the task with much less search, and directly matching its distribution transfers this short-search behavior to the student.

The RLSD/SRPO adaptation instead uses the token-level self-distillation signal as advantage weights on incorrect trajectories. Its average trajectory length recovers to 35.5 turns, close to GRPO's 36.1 and substantially above the direct OPSD adaptation's 30.9, while its accuracy remains 11.8. This recovery supports using the teacher signal to modulate the update magnitude rather than as a direct optimization target; however, the unchanged accuracy also shows that token-level weighting alone is insufficient in this setting.

\begin{table}[H]
\centering
\small
\setlength{\tabcolsep}{4pt}
\resizebox{\textwidth}{!}{%
\begin{tabular}{l c c c c c}
  \toprule
  \textbf{Method} & \textbf{Evidence Anchors} & \textbf{Incorrect feedback} & \textbf{BC-Sub} & \makecell{\textbf{Sign agreement} \\ \textbf{with full setting (\%)}} & \makecell{\textbf{Final-step penalty} \\ \textbf{amplified (\%)}} \\
  \midrule
  GRPO & No & No & 12.8 & -- & -- \\
  SSPO w/o Evidence Anchors & No & Yes & 12.4 & 32.8 & 92.4 \\
  SSPO w/o Incorrect Feedback & Yes & No & 14.0 & 81.1 & 46.7 \\
  \textbf{Ours (SSPO)} & Yes & Yes & \textbf{14.5} & 100.0 & 86.3 \\
  \bottomrule
\end{tabular}}
\caption{Ablation of the two privileged-information sources. BC-Sub results are measured after 50 training steps. The two rightmost columns are computed by rescoring 512 incorrect trajectories, using the sign under the full setting as a reference. ``Final-step penalty amplified'' denotes the proportion of trajectories for which the final-step weight is greater than one.}
\label{tab:privileged-ablation}
\end{table}

\begin{figure}[!ht]
    \centering
    \begin{minipage}{0.46\linewidth}
        \centering
        \includegraphics[width=\linewidth]{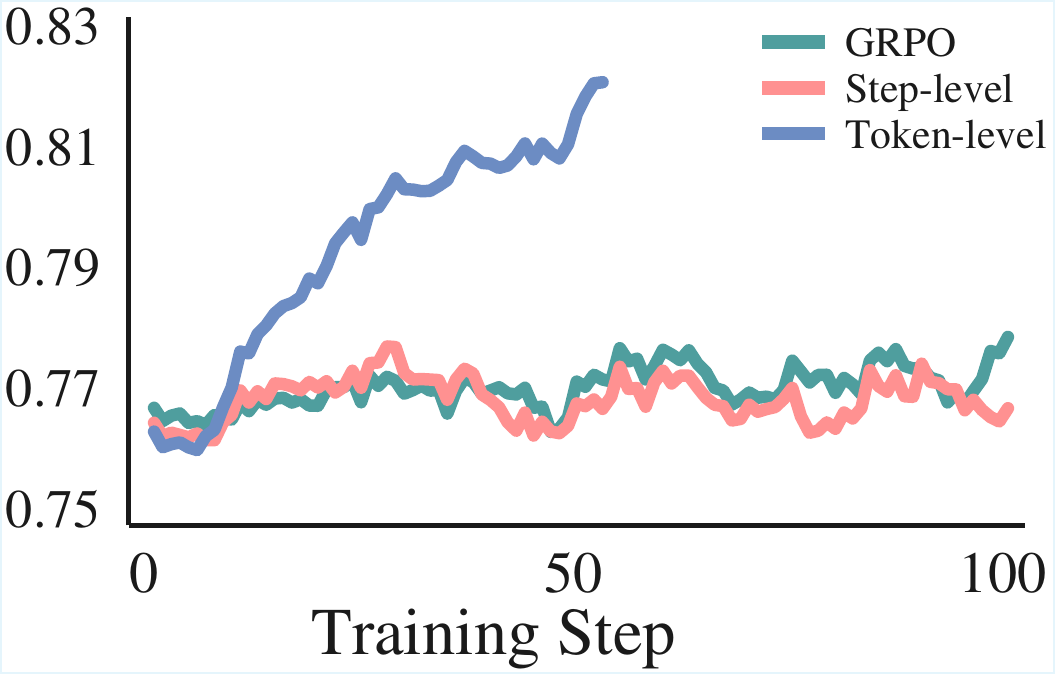}
        \caption{Entropy curves of GRPO, step-level, and token-level supervision during training.}
        \label{fig:entropy-compare}
    \end{minipage}
    \hfill
    \begin{minipage}{0.5\linewidth}
        \centering
        \includegraphics[width=\linewidth]{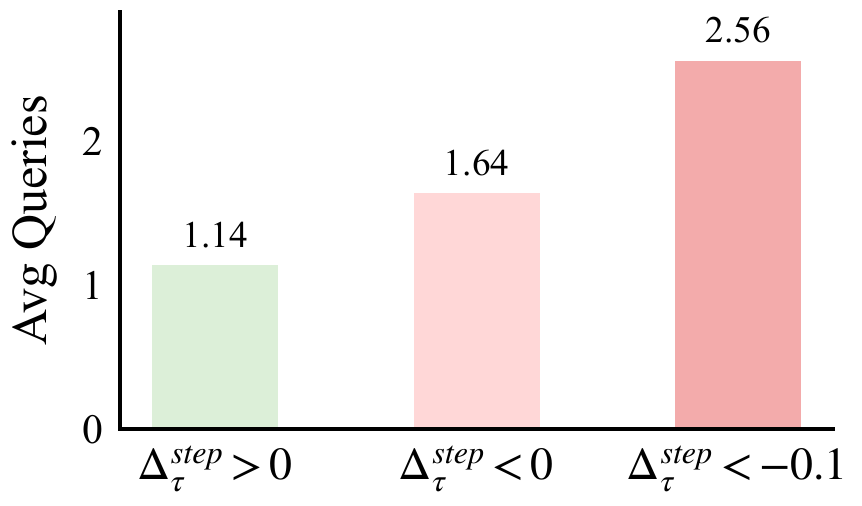}
        \caption{Average number of queries under different step-level privileged-information gains.}
        \label{fig:queries-gain}
    \end{minipage}
\end{figure}

\paragraph{Step-Level Rather Than Token-Level.}
We then isolate supervision granularity by holding advantage weighting, Evidence Anchors, and incorrect-trajectory routing fixed. The token-level RLSD/SRPO adaptation assigns a separate weight to every token (Appendix~\ref{app:token-level}), whereas SSPO shares one joint-probability weight across the reasoning and tool-calling tokens that form a complete information-seeking step. This controlled change improves BC-Sub accuracy from 11.8 to 14.5 while maintaining the average trajectory length (35.5 vs. 37.5 turns). The entropy curves in Figure~\ref{fig:entropy-compare} provide complementary evidence: token-level weighting produces steadily increasing entropy, whereas step-level weighting remains more stable. These results support aligning the supervision granularity with the complete search action rather than fragmenting it across individual tokens.

\paragraph{Roles of Evidence Anchors and Incorrect-Answer Feedback.}
Table~\ref{tab:privileged-ablation} separates the two sources of privileged information. Incorrect-answer feedback alone does not improve over GRPO (12.4 vs. 12.8), whereas Evidence Anchors alone improve accuracy to 14.0. Combining both yields the best result, 14.5, showing that Evidence Anchors provide the primary process-supervision signal and incorrect-answer feedback is complementary.

To examine how they contribute to the full step score, we rescore the same 512 incorrect trajectories under each single-information setting. This is a decomposition of the deployed signal rather than an assumption that the full setting is an unbiased ground-truth label. Evidence Anchors alone agree with the sign of the full signal on 81.1\% of steps, compared with 32.8\% for incorrect-answer feedback alone, indicating that Evidence Anchors primarily determine the evaluation of intermediate information-seeking actions. Conversely, incorrect-answer feedback amplifies the final-step penalty in 92.4\% of trajectories, compared with 46.7\% under Evidence Anchors alone. Thus, the incorrect answer primarily identifies the failed terminal conclusion, while Evidence Anchors ground credit assignment across the preceding search process.

\section{Case Study: Which Steps are Exempted from Penalty?}

To better understand why SSPO is effective, we analyze how the teacher assigns penalties across different types of steps.
We find that \textbf{the teacher does not simply reward steps based on their lexical overlap with evidence anchors. Instead, it modulates penalties according to whether a step performs a focused and informative verification action}.

Specifically, steps that receive reduced penalties are typically compact, well-scoped queries targeting a single local constraint implied by the evidence anchors, as shown in Figure~\ref{fig:pos-step}. Although these steps may contain fewer anchor words, they effectively reduce uncertainty along the reasoning chain.
In contrast, steps that receive stronger penalties are often long, diffuse queries that concatenate multiple surface terms from the question or anchors without forming a precise information need, as shown in Figure~\ref{fig:neg-step}.
Despite their high lexical overlap with anchors, such steps receive amplified penalties, indicating that the teacher prioritizes information quality over word matching.
This behavior is further supported by quantitative analysis. 
Figure~\ref{fig:queries-gain} shows the relationship between the average number of queries in search tool calls and the privileged-information gain. Steps with positive privileged-information gain $\Delta_\tau^{step} > 0$ use significantly fewer queries than those with negative gain $\Delta_\tau^{step} < 0$.
Overall, SSPO encourages targeted, evidence-grounded information-seeking by reducing penalties for precise verification steps while amplifying penalties for broad and unfocused exploration.

\begin{figure}[!ht]
\begin{tcolorbox}[
    colback=green!5,
    colframe=green!45!black,
    fonttitle=\bfseries,
    boxrule=0.8pt,
    arc=2pt
]
\begin{lstlisting}[style=promptstyle, backgroundcolor=\color{green!5}]
<think>
Perhaps the specific ERNI employee referenced is not the person listed in the program committee (Florian, Dominik etc). The above LinkedIn snippet for Patrick Pfeifer shows "iSAQB - Foundation Level (CPSA-F)", which matches the certification.
Now does Patrick Pfeifer specialize in medical technology software? Potentially, as ERNI works in medtech, and "Patrick Pfeifer" might be focusing on medical solutions, but we need verification.
Search "Patrick Pfeifer ERNI medical technology".
</think>
<tool_call>
"search queries": ["Patrick Pfeifer ERNI medical technology"]
</tool_call>
\end{lstlisting}
\end{tcolorbox}
\caption{$\Delta_\tau^{step} > 0$: compact single-query verification targeting one specific entity-attribute relation.}
\label{fig:pos-step}
\end{figure}

\begin{figure}[!ht]
\begin{tcolorbox}[
    colback=red!4,
    colframe=red!45,
    fonttitle=\bfseries,
    boxrule=0.8pt,
    arc=2pt
]
\begin{lstlisting}[style=promptstyle, backgroundcolor=\color{red!4}]
<think>
ERNI matches the consultancy with offices across Europe and Asia.
Now find an ERNI employee who is a "medical technology software specialist" and holds a "foundational certification in software architecture". The foundation certification could refer to the iSAQB CPSA-F certification.
We also have that this employee serves on the conference committee for BaselOne and recently co-founded an initiative linking local industry and academia.
The program committee includes Ix-chel Ruiz, Dominik Berger, Marcus Fihlon, Robin Mezger, Florian Stampe, Frank Baier, etc. Among these, perhaps one is a medical technology software specialist.
But we could also search for "BaselOne conference committee ERNI".
</think>
<tool_call>
"search queries": [
  "BaselOne conference committee ERNI",
  "ERNI medical technology software architect BaselOne",
  "ERNI medical technology software specialist BaselOne committee",
  "CPSA Foundation ERNI employee"]
</tool_call>
\end{lstlisting}
\end{tcolorbox}
\caption{$\Delta_\tau^{step} < 0$: diffuse multi-query search with high lexical overlap but poor localization.}
\label{fig:neg-step}
\end{figure}

\section{Conclusion}

We presented SSPO, a step-level self-distillation method that extends On-Policy Self-Distillation from single-turn reasoning to multi-turn deep search agents. To construct a meaningful self-teacher in a setting where neither reference solutions nor execution feedback are available, we introduced Evidence Anchors—structured, step-aligned privileged information whose granularity matches that of the search process itself. From this teacher, SSPO derives a self-distillation signal, recasts it as step-level advantage weights, and applies it only to incorrect trajectories, thereby injecting fine-grained process supervision while sidestepping both privileged-information leakage and the ambiguous gradients that direct distillation would otherwise impose on already-correct rollouts.

Across BrowseComp, GAIA, and FRAMES, SSPO surpasses GRPO trained for twice as many gradient steps, while sustaining higher rewards and more stable gradient norms throughout optimization. Beyond the aggregate gains, our analysis reveals a coherent shaping of the policy update: penalties contract on compact, evidence-grounded verification steps and expand on diffuse, underspecified queries. This suggests that step-level self-distillation guides the agent toward more targeted and efficient information-seeking, rather than simply lifting final-answer accuracy.

\bibliographystyle{plain}
\bibliography{ref}

%%%%%%%%%%%%%%%%%%%%%%%%%%%%%%%%%%%%%%%%%%%%%%%%%%%%%%%%%%%%

\newpage

\appendix

\section{Related Work}

\subsection{Reinforcement Learning with Verifiable Rewards (RLVR)}

Building on works such as DeepSeek-MATH~\cite{shao2024deepseekmathpushinglimitsmathematical} and DeepSeek-R1~\cite{Guo_2025}, RLVR has become a standard component in LLM training pipelines and is widely regarded as essential for improving reasoning capabilities across domains, including but not limited to mathematics~\cite{zeng2025simplerlzooinvestigatingtamingzero,yu2025dapoopensourcellmreinforcement}, logical reasoning~\cite{xie2025logicrlunleashingllmreasoning,liu2025synlogicsynthesizingverifiablereasoning,wu2025miragemethodmodeltaskalignment}, coding~\cite{code-r1,wu2025recodeupdatingcodeapi}, and verifiable agent tasks~\cite{jin2025empirical,jin2025search,zhang2025toolr1sampleefficientreinforcementlearning}.
Although RLVR is considerably more complex to implement than SFT, its On-Policy nature confers a qualitatively distinct advantage: by exploring the environment during training, the model is not confined to imitating fixed teacher trajectories and can autonomously discover effective reasoning strategies. 
DeepSeek-R1~\cite{Guo_2025} demonstrates that, without any process-level supervision, RL training can spontaneously elicit sophisticated behaviors such as backtracking and self-reflection—phenomena collectively referred to as ``aha moments.'' 
This stands in contrast to SFT, which by directly fitting a static teacher distribution fundamentally forecloses such exploration~\cite{schmidt-2019-generalization,pmlr-v267-chu25c,wu2026generalizationsftreinforcementlearning,shenfeld2026selfdistillationenablescontinuallearning}.

\subsection{Process Rewards in Search Agent Training}

To address the sparsity of outcome-only rewards in long-horizon agent settings, several recent works introduce process rewards that evaluate intermediate steps during search. 
CriticSearch~\cite{zhang2025criticsearchfinegrainedcreditassignment} employs a frozen asymmetric critic LLM that retrospectively evaluates each interaction turn using privileged information from the complete trajectory and gold answers, converting these assessments into dense turn-level rewards for policy optimization. 
PPR~\cite{xu2026principle} trains a dedicated principle-based process reward model that grounds step-wise judgments in interpretable principles such as correctness, relevance, and consistency, and further introduces a reward normalization strategy to balance local process fidelity against global task success. 
SmartSearch~\cite{wen2026smartsearchprocessrewardguidedquery} takes a query-centric view, designing a dual-level credit assessment mechanism that scores each intermediate search query for both novelty and usefulness, and uses these scores to selectively refine low-quality queries via a separately trained smaller model.

While these approaches demonstrate the value of fine-grained supervision for search agents, they share two notable limitations. First, their experimental benchmarks are primarily composed of standard multi-hop QA tasks, such as HotpotQA~\cite{yang2018hotpotqadatasetdiverseexplainable} and 2WikiMultiHop~\cite{xanh2020_2wikimultihop}, which involve relatively shallow retrieval chains. 
None of them evaluate on BrowseComp-style tasks that require navigating dozens of search and browse steps to resolve highly constrained, multi-conditional queries—precisely the setting where per-step credit assignment is most critical. 
Second, all three methods rely on a separate LLM as a per-step evaluator, which introduces additional inference cost and a dependency on the quality and calibration of that external scorer. 
In contrast, SSPO eliminates the need for an explicit step-level reward model by converting self-distillation signals directly into step-level advantage weights, making fine-grained process supervision both practically lightweight and tightly integrated with the on-policy training objective.

\subsection{On-Policy (Self-)Distillation}

On-Policy Distillation (OPD)~\cite{lu2025onpolicydistillation,song2026surveyonpolicydistillationlarge,li2026rethinking,xu2026tiptokenimportanceonpolicy} improves upon traditional off-policy supervised fine-tuning (SFT) by allowing the student to generate its own rollouts while using teacher logits for supervision, thereby reducing the distribution mismatch between training and inference. 
However, OPD still relies on a teacher model that operates in a compatible vocabulary space and is strictly more capable than the student, which limits its practical applicability.

A growing body of work seeks to remove this dependency by constructing self-teachers from privileged information or environmental feedback. 
Self-Distilled Reasoner~\cite{zhao2026selfdistilledreasoneronpolicyselfdistillation} shows that On-Policy Self-Distillation (OPSD), when augmented with reference solutions as privileged prefixes, achieves strong performance on single-turn mathematical reasoning tasks. 
RL via Self-Distillation~\cite{hubotter2026reinforcement} extends this paradigm to tool-use and coding scenarios, where environment feedback—such as compiler errors or execution outputs—serves as a form of privileged supervision. 
Self-Distilled RLVR (RLSD)~\cite{yang2026selfdistilledrlvr} further identifies a critical limitation of direct OPSD: optimizing the student toward a teacher conditioned on privileged information can introduce information leakage, where the policy implicitly exploits signals unavailable at test time. 
To address this issue, RLSD~\cite{yang2026selfdistilledrlvr} proposes converting distillation signals into advantage weights rather than directly using them as gradient targets, thereby preserving the on-policy reward as the primary optimization objective. 
SRPO~\cite{li2026unifyinggrouprelativeselfdistillationpolicy} further observes that applying self-distillation uniformly across both correct and incorrect trajectories introduces ambiguous optimization signals, and instead advocates restricting OPSD signals to incorrect trajectories. 
CRISP~\cite{sang2026crispcompressedreasoningiterative} explores iterative self-policy distillation for compressing reasoning chains, while On-Policy Context Distillation~\cite{ye2026onpolicycontextdistillationlanguage} studies distillation from context-augmented teachers in language model settings. 
More broadly, self-distillation has also been shown to support continual learning without catastrophic forgetting~\cite{shenfeld2026selfdistillationenablescontinuallearning}.

Our work builds on these insights and extends OPSD to multi-turn deep search agents, a setting that introduces two key challenges not addressed by prior work: 
(1) how to construct meaningful privileged information for open-ended information retrieval tasks, where neither reference solutions nor execution feedback are directly available; and 
(2) how to define an appropriate supervision granularity that aligns with the natural unit of search behavior — information-seeking actions — rather than defaulting to token-level signals.

\section{More Experimental Details}
\label{app:more-exp-detail}

\paragraph{Scaffold Details.}
Our scaffold is based on the ReAct paradigm~\cite{yao2023reactsynergizingreasoningacting}, enabling interaction with two tools: \texttt{search} and \texttt{browse} (see Appendix~\ref{app:tool-schema}). We use no specialized system prompts during either training or evaluation, providing only the necessary tool descriptions. Except for the teacher model, the user prompt contains only the question itself.
The model is required to invoke a tool at each step, except for the final step where it outputs the answer, until termination or reaching the step limit. If no valid tool call is produced at any step, the trajectory is terminated and scored according to the final response. Following prior work~\cite{liu2025webexplorer,li2025websailornavigatingsuperhumanreasoning}, we use an LLM judge to evaluate answer correctness.

\paragraph{Training Details.}
We use Qwen3-8B~\cite{yang2025qwen3technicalreport} as the base model for all experiments.
For cold-start Supervised Fine-Tuning (SFT), we set the batch size to 32 and the learning rate to $1\times10^{-5}$, with linear warmup followed by cosine decay, and train the model for 1k steps.
For On-Policy learning, we train on approximately 6k samples using a fixed learning rate of $1\times10^{-6}$ and a batch size of 64, with 8 rollouts per question. For our method, we set $\epsilon = 0.2$, and reinitialize the teacher model with the current policy model every 50 training steps.
Throughout training, we set the maximum context length to 128K and the maximum number of agent steps to 100. 

\paragraph{Loss Masking.}
For both SFT and On-Policy learning, we compute the training loss only on agent-generated tokens, including the Thought and Action segments, as well as the final answer when applicable. 
Tokens returned by the environment, such as search results and browsed page contents, are provided only as context for subsequent agent decisions and are masked out from the loss. 
This is because observation tokens are generated by external tools rather than by the agent policy, and therefore should not be predicted or optimized as part of the agent’s action distribution.

\section{Agent Tool Schemas}
\label{app:tool-schema}

Similar to WebExplorer~\cite{liu2025webexplorer}, we provide the LLM with two tools for information retrieval:

\paragraph{Search Tool.}
Our search tool enables the LLM to issue multi-keyword queries simultaneously. It leverages the Serper API\footnote{https://serper.dev/} to return relevant information triplets \texttt{(title, URL, snippet)} to the LLM.

\begin{tcolorbox}[
  title=Search Tool Schema,
  fonttitle=\bfseries\small,
  colback=myblue,
  colframe=myframe,
  colbacktitle=mytitle,
  coltitle=black,
  boxrule=0.5pt,
  arc=4pt
]
\begin{verbatim}
type: function
function:
  name: search
  description: Web search.
  parameters:
    type: object
    properties:
      queries:
        type: array
        description:
          The queries will be sent to Google via Serper API. You will 
          get the brief search results with (title, url, snippet)s for 
          each query.
        items:
          type: string
    required: queries
\end{verbatim}
\end{tcolorbox}

\paragraph{Browse Tool.}
Our browse tool retrieves and processes content from specific URLs using content extraction and language model capabilities. Specifically, content extraction is powered by Jina’s service\footnote{https://jina.ai/}, while long-context retrieval is handled by another LLM with a context window exceeding 256k tokens.

\begin{tcolorbox}[
  title=Browse Tool Schema,
  fonttitle=\bfseries\small,
  colback=myblue,
  colframe=myframe,
  colbacktitle=mytitle,
  coltitle=black,
  boxrule=0.5pt,
  arc=4pt
]
\begin{verbatim}
type: function
function:
  name: browse
  description: Explore specific information in a url.
  parameters:
    type: object
    properties:
      url:
        type: string
        description:
          The url will be browsed, and the content will be sent to a 
          Large Language Model (LLM) as the based information to 
          answer a query.
      query:
        type: string
        description:
          The query to this url content.
    required: [url, query]
\end{verbatim}
\end{tcolorbox}

\section{Cold-Start Trajectory Collection}
\label{app:cold-start}

\subsection{QA Generation}

To obtain high-quality teacher trajectories for cold-start initialization, we first require sufficiently challenging QA pairs~\cite{li2025websailornavigatingsuperhumanreasoning,li2025websailorv2bridgingchasmproprietary,tao2025webshaperagenticallydatasynthesizing,liu2025webexplorer,li2025webthinkerempoweringlargereasoning}.
However, for a long time, such data has been largely lacking in the open-source community.
WebExplorer~\cite{liu2025webexplorer} proposes a model-based approach for QA pair synthesis, enabling the generation of sufficiently challenging examples without constructing large-scale web topology graphs.
It divides QA pair synthesis into two stages: \textit{Model-Based Exploration} and \textit{Iterative Query Evolution}, and employs the costly Claude model in both stages.
When using other, lower-cost API models, the difficulty of the generated questions tends to drop significantly, as shown in Table~\ref{tab:qa-generate-diff-models}.

\begin{table}[!ht]
    \centering
    \begin{tabular}{l|cccc}
        \toprule
        \textbf{Model} & \textbf{Accuracy} & \textbf{\#Tool Calling} & \textbf{\#Search} & \textbf{\#Browse} \\
        \midrule
        Claude-4-Sonnet~\cite{anthropicIntroducingClaude} & \textbf{58.3} & \textbf{20.2} & \textbf{15.0} & \textbf{5.2} \\
        \midrule
        Grok-4.1-Fast~\cite{grok41fast} & 81.7 & 12.5 & 8.0 & 4.5 \\
        GLM-4.6~\cite{glm46} & 75.5 & 12.5 & 8.2 & 4.3 \\
        DeepSeek-V3.2~\cite{deepseekai2025deepseekv32pushingfrontieropen} & 80.0 & 8.9 & 6.2 & 2.7 \\
        \midrule
        DS-V3.2 + Modified Prompt & 62.0 & 15.0 & 11.1 & 3.9 \\
        \bottomrule
    \end{tabular}
    \caption{Comparison of question difficulty across different models, evaluated by accuracy and the average number of tool calls (measured on GPT-5-Nano~\cite{gpt5nano}).}
    \label{tab:qa-generate-diff-models}
\end{table}

We find that the prompt used in the second stage is somewhat too simplistic, making it difficult for models with weaker instruction-following ability to effectively increase the difficulty of the generated questions.
We revise it to the version shown in Figure~\ref{fig:evolution-prompt-template}, which substantially improves question difficulty even when using DeepSeek-V3.2~\cite{deepseekai2025deepseekv32pushingfrontieropen} under the same setting.
As shown in Table~\ref{tab:qa-generate-diff-models}, on 100 generated instances, the accuracy of GPT-5-Nano~\cite{gpt5nano} decreases from 80.0 to 62.0, while the average number of tool calls increases from 8.9 to 15.
We then use the improved pipeline with the lower-cost DeepSeek-V3.2~\cite{deepseekai2025deepseekv32pushingfrontieropen} to generate over 6,000 QA pairs.

\begin{figure*}[!ht]
\centering
\begin{minipage}{\textwidth}
\begin{promptbox}{}
\begin{lstlisting}[style=promptstyle]
You must **significantly increase** the difficulty of the following question while ensuring the correct answer remains **uniquely identifiable**.

Original question: {orig_question}
Original truth: {truth}

You **must aggressively** apply ALL of the following strategies to make the question harder:
1. **Remove highly specific clues**: delete dates, numbers, full names, institutions, awards, locations, publication names, etc.; keep only the minimal signals needed for unique identification
2. **Blur and generalize**: replace concrete information with uncertain or approximate descriptions, while keeping the truth uniquely resolvable
3. **Add distractor-like similar entities**: introduce misleading cues so shallow reasoning fails and deeper inference is required
4. **Refer indirectly to the entity**: use less-common identifiers---predecessor/successor relationships, indirect influence, obscure nicknames, abstract impact, associated figures, etc.
5. **Iterative escalation**: perform **5 evolution steps**; each step must remove or obscure at least one previously clear attribute, and be **strictly harder** than the previous version
6. Ensure **uniqueness of the final truth**: despite the ambiguity, the question must still map to the exact same truth

You may use search and browsing tools to verify uniqueness during rewriting.

---

Output format:

For each iteration:
<question>
{more difficult question version}
</question>

After the 5th iteration, output the final result:
<answer>
<question>{the most difficult version}</question>
<truth>{the exact same truth}</truth>
</answer>

Do NOT include any other explanations, comments, or formats.
If any iteration fails to increase difficulty, the task is considered failed.
\end{lstlisting}
\end{promptbox}
\end{minipage}
\caption{Modified Prompt template used in the \textit{Iterative Query Evolution} stage.}
\label{fig:evolution-prompt-template}
\end{figure*}

\subsection{Teacher Trajectories}

With high-quality QA pairs in hand, we then employ a teacher model to generate ReAct-style reasoning trajectories~\cite{yao2023reactsynergizingreasoningacting}.
We deploy GPT-OSS-120B~\cite{gptoss} as the teacher model and set the reasoning effort to high.
This setup yields the model’s native long-form reasoning trajectories, rather than user-facing compressed CoTs~\cite{wei2023chainofthoughtpromptingelicitsreasoning}.
In addition, when comparing trajectories from GPT-OSS-120B~\cite{gptoss} with those from other commercial flagship models, we observe two key characteristics:
(1) Compared to flagship models, GPT-OSS-120B~\cite{gptoss} makes more tool calls.
(2) When both browse and search tools are available, GPT-OSS-120B~\cite{gptoss} invokes browse to retrieve webpage content significantly more frequently than other models.

\begin{table}[!ht]
    \centering
    \begin{tabular}{l|cccc}
        \toprule
        \textbf{Model} & \textbf{\#Tool Calling} & \textbf{\#Search} & \textbf{\#Browse} & \textbf{\#Browse Frac} \\
        \midrule
        Claude-4-Sonnet~\cite{anthropicIntroducingClaude} & 5.67 & 4.48 & 1.19 & 21.0\% \\
        GLM-4.6~\cite{glm46} & 4.92 & 3.87 & 1.05 & 21.3\% \\
        \midrule
        GPT-OSS-120B~\cite{gptoss} & \textbf{18.14} & 11.90 & 6.24 & \textbf{34.4\%} \\
        \bottomrule
    \end{tabular}
    \caption{Comparison of Tool Usage Across Different Models.}
    \label{tab:cold-start-tool}
\end{table}

As shown in Table~\ref{tab:cold-start-tool}, for the same queries, GPT-OSS-120B~\cite{gptoss} performs more than three times as many tool calls as other models and more frequently invokes the browse tool to retrieve detailed information from the web.
This highlights differences in reasoning strategies across models: flagship models often narrow down candidate answers to a very small set using internal knowledge in the first step, requiring only minimal verification.
In contrast, GPT-OSS-120B~\cite{gptoss} does not exhibit similarly rich internal knowledge in domains such as the social sciences and arts~\cite{bi2025gptossgoodcomprehensiveevaluation}, making its trajectories more suitable for student models that lack sufficient internal knowledge.
Although we use only 4,000 trajectories—significantly fewer than the 13,000 used in WebExplorer~\cite{liu2025webexplorer}, we achieve performance on par with, or even surpassing, WebExplorer-8B-SFT~\cite{liu2025webexplorer}. 
This highlights the advantage of GPT-OSS-120B~\cite{gptoss} as a teacher model.

Table~\ref{tab:sft-data} presents the statistics of the final teacher trajectories.
We ultimately retain only correct trajectories and remove those with obvious tool-calling errors.
In addition, we observe that incorrect trajectories are approximately three times longer than correct ones, underscoring both the necessity and urgency of fine-grained supervision for erroneous trajectories.

\begin{table}[!ht]
    \centering
    \begin{tabular}{c|cccc}
    \toprule
     & Accuracy & \textbf{\#Tool} & \textbf{\#Browse} & \textbf{\#Search} \\
    \midrule
    All & 0.74 & 32.1 & 10.5 & 21.6 \\
    Correct & - & 20.5 & 8.3 & 12.2 \\
    Incorrect & - & 65.5 & 16.9 & 48.6 \\
    \bottomrule    
    \end{tabular}
    \caption{Statistics of the teacher trajectories.}
    \label{tab:sft-data}
\end{table}

\section{More Experimental Results}
\label{app:more-exp-res}

\begin{figure}[!ht]
  \centering
  \includegraphics[width=\textwidth]{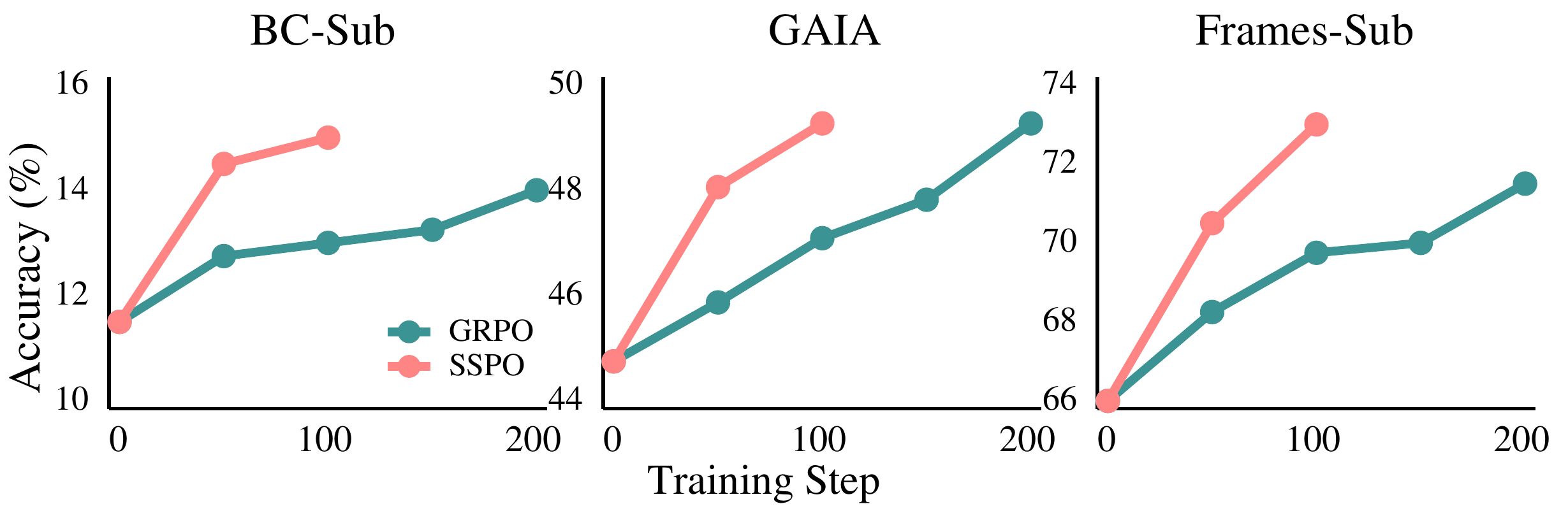}
  \caption{Training dynamics of GRPO and SSPO on BC-Sub, GAIA, and Frames-Sub. }
  \label{fig:main-res}
  
\end{figure}

\paragraph{Training Dynamics.}
As shown in Figure~\ref{fig:main-res}, SSPO consistently outperforms GRPO across all three benchmarks throughout training, and achieves stronger performance with fewer training steps, indicating both higher sample efficiency and better final performance.

\paragraph{Acceptable Training Overhead.}
Although SSPO achieves significant performance gains, it requires one additional forward pass to obtain the teacher logits. 
When training on two 8-accelerator nodes with 140 GB memory each, the time breakdown across different stages of each training step is reported in Table~\ref{tab:time-proportion}.
The additional teacher forward pass accounts for only $\sim$5\% of the total step time, making the overhead negligible relative to the performance gains.

\begin{table}[!ht]
    \centering
    \begin{tabular}{c|ccccc}
    \toprule
     & Collect Traj & Update Actor & Policy logP & Teacher logP & Other\\
    \midrule
    Proportion (\%) & 59.2 & 23.3 & 5.5 & 6.5 & 5.5 \\
    \bottomrule    
    \end{tabular}
    \caption{The proportion of time consumed by different components within each training step.}
    \label{tab:time-proportion}
\end{table}

\section{More Details about Ablation Study}

\subsection{Training Objective of OPSD}
\label{app:opsd-obj}

We replace the advantage $A_t^{(i)}$ in Equation~\ref{eq:grpo} with $\log P_T(y\mid c_{privileged},q,y_{<t}) - \log P_S(y\mid q,y_{<t})$ as the training objective of OPSD.
This objective is more tractable than computing the KL divergence over the full vocabulary and has been widely shown to be effective~\cite{lu2025onpolicydistillation}.

\subsection{Token-Level Advantage Weights}
\label{app:token-level}

Similar to Step-Level Self-Distilled Advantage Weights in Section~\ref{sec:ssaw}, we can compute the privileged-information gain for each token $y_t \in y^{(i)}$ as:
\begin{equation}
  \Delta_t^{token} = \text{\texttt{sg}}(\log P_T(y_t,|c_{privileged},q,y_{<t}) - \log P_S(y_t|q,y_{<t}))
\end{equation}
We can then compute the advantage weight for each token:
\begin{equation}
  w_t = \min(\exp(\text{sign}(A^{(i)})\cdot\Delta_t^{token}),1+\epsilon)
\end{equation}
Finally, we replace the advantage term with $\hat{A}^{(i)}_t = w_tA_t^{(i)}\text{ \texttt{if} }R_{final}<1\text{ \texttt{else} }A^{(i)}$.

\section{More Details about Evidence Anchors}
\label{app:evidence-anchors}

\paragraph{Evidence Anchor Quality Validation.} 
As shown in Figure~\ref{fig:evidence-anchors-prompt-template}, during Evidence Anchor collection, we also require the model to provide the URL of each source page. 
Although URLs themselves contain limited instructional semantic information and are therefore not included in the teacher prefix, we use them for automatic quality validation.
Specifically, to reduce the risk that the LLM fabricates non-existent sources, we use Jina to access each provided webpage and verify both URL accessibility and whether the retrieved page title matches the source title reported by the LLM. 
Encouragingly, thanks to recent improvements in LLM capability and our use of \textit{search} and \textit{browse} tools during anchor collection, only a very small fraction of Evidence Anchors contain inaccessible URLs. 
We filter out these invalid anchors and remove QA pairs containing invalid Evidence Anchors from the training set.

\begin{figure}[!ht]
  \centering
  \includegraphics[width=0.98\textwidth]{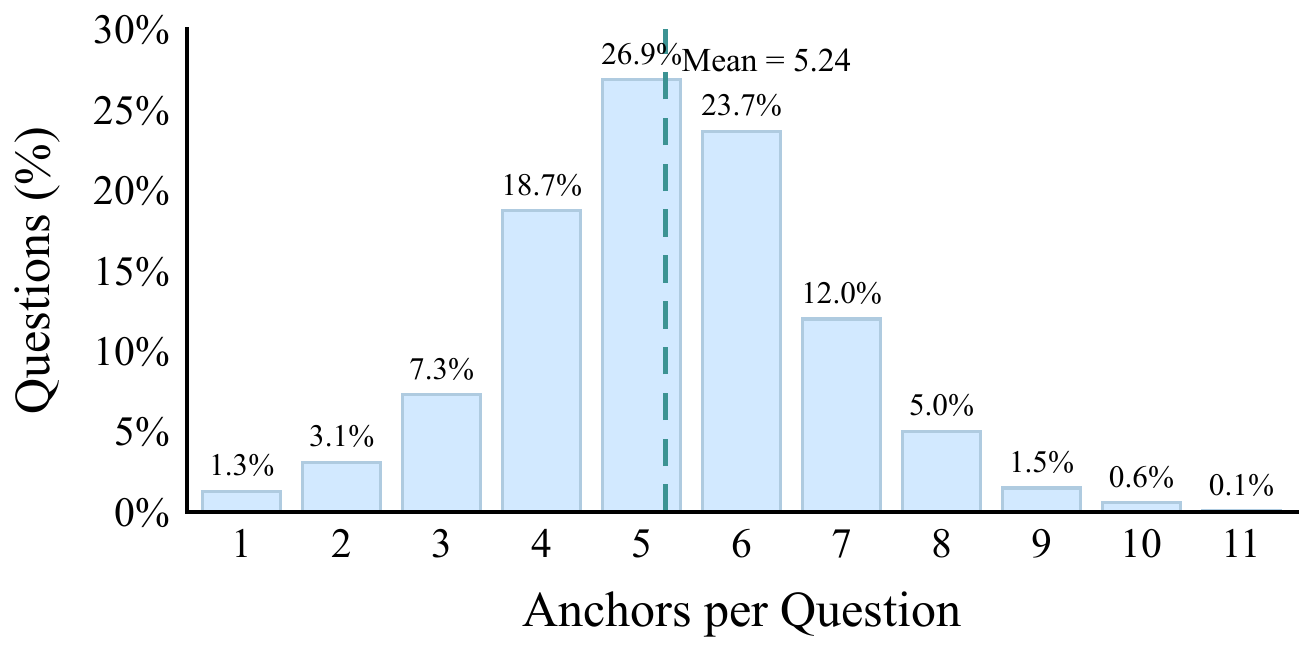}
  \caption{Statistics of Evidence Anchor Counts.}
  \label{fig:evidence-anchors-cnt}
\end{figure}

\paragraph{Evidence Anchor Statistics.}
Figure~\ref{fig:evidence-anchors-cnt} shows the distribution of the number of evidence anchors per question in our training data.
The distribution is centered around 5 anchors per question, with a mean of 5.24.
Most questions contain between 4 and 6 anchors, which together account for the majority of the dataset.
The frequency drops off on both sides, with very few questions having fewer than 3 or more than 8 anchors.
This indicates that our data construction process produces moderately sized evidence sets, balancing coverage and conciseness for effective supervision.

\begin{figure*}[!ht]
\centering
\begin{minipage}{\textwidth}
\begin{promptbox}{}
\begin{lstlisting}[style=promptstyle]
CRITICAL ROLE: You are a strict Verification Engine. Your sole mission is to find evidence that the [Standard Answer] SPECIFICALLY satisfies each constraint of the [Question].
STRATEGIC SEARCH MANDATE:
1. DECOMPOSE: Break the [Question] into independent, verifiable conditions.
2. ANCHORING SEARCH: For each condition, your search query MUST include the [Standard Answer] or be related to the [Standard Answer].
   - BAD Search: "who is the COOP leader in Amarillo" (This is solving)
   - GOOD Search: "Angela Margrave National Weather Service Amarillo COOP leader" (This is verifying)
3. PROVE SATISFACTION: A condition is only 'Verified' if you find a source that explicitly links the [Standard Answer] to that specific requirement.
OUTPUT FORMAT (Strict XML):
<evidences>
  <evidence>
    <condition>The specific requirement/condition extracted from the [Question]</condition>
    <source_title>The title of the source web page or article</source_title>
    <source_url>The full URL of the supporting web page</source_url>
    <explanation>Briefly explain how this evidence explicitly proves that the [Standard Answer] satisfies this specific condition</explanation>
  </evidence>
</evidences>
[Example]:
Question: A U.S. citizen science initiative relies on volunteers to collect daily meteorological data. The data gathered is vital for a federal agency operating within the same department as the body responsible for mapping the ocean floor. The forecast office for a region known for a major city with a famous public art installation of partially buried cars has a specific leader for this program. What is the name of this program leader?
Answer: Angela Margrave

Your Output:
<evidences>
  <evidence>
    <condition>The forecast office must be located in a region known for a major city with a famous public art installation of partially buried cars.</condition>
    <source_title>Cadillac Ranch - Wikipedia</source_title>
    <source_url>https://en.wikipedia.org/wiki/Cadillac_Ranch</source_url>
    <explanation>The evidence shows that Cadillac Ranch is a famous art installation of half-buried cars located in Amarillo, Texas. This proves the required forecast office region is Amarillo.</explanation>
  </evidence>
  <evidence>
    <condition>The specific leader for this meteorological program at the identified forecast office (Amarillo) must match the answer.</condition>
    <source_title>National Weather Service - COOP Recruitment Amarillo</source_title>
    <source_url>https://www.weather.gov/ama/COOP_Recruitment</source_url>
    <explanation>The official NWS page explicitly lists Angela Margrave as the COOP Program Leader at the Amarillo, TX forecast office, confirming the standard answer perfectly satisfies the final requirement.</explanation>
  </evidence>
</evidences>

[Question]: {question}
[Answer]: {answer}
\end{lstlisting}
\end{promptbox}
\end{minipage}
\caption{Prompt template used to collect Evidence Anchors.}
\label{fig:evidence-anchors-prompt-template}
\end{figure*}

\section{Limitations and Future Work}

Due to the high cost associated with API usage (e.g., Serper, Jina, and LLM services), the scale of both data construction and experimental evaluation is constrained. 
For both cold-start and on-policy training, we rely on only a few thousand samples. 
Moreover, due to computational and time limitations, our experiments are restricted to 8B-scale models, and we do not evaluate performance on larger model sizes.
Additionally, due to the English-only chain-of-thought characteristics of the teacher model (GPT-OSS-120B~\cite{gptoss}), both training and evaluation are confined to English, without incorporating multilingual data such as Chinese or Japanese. 
This may limit the generality of our findings to broader multilingual settings.

In future work, we aim to investigate the role of fine-grained supervision in improving existing RLVR methods across more diverse agent scenarios. 
We also plan to scale up both the data and experimental scope, including exploring larger model sizes and extending the framework to multilingual settings.

% \section{Broader Impacts}

% This work aims to improve the reliability of deep search agents by encouraging more precise and evidence-grounded information-seeking behavior. Such improvements may benefit applications including scientific research, complex question answering, and fact verification, where accurate retrieval and reasoning are important.

% At the same time, more capable search agents may introduce risks, such as misuse for large-scale automated information gathering or over-reliance on generated outputs, which may propagate errors. To mitigate these risks, we emphasize the importance of evidence grounding, source attribution, and human verification in downstream applications.

%%%%%%%%%%%%%%%%%%%%%%%%%%%%%%%%%%%%%%%%%%%%%%%%%%%%%%%%%%%%
% \clearpage
% \input{checklist.tex}

\end{document}